# Semantic interpretation for convolutional neural networks: What makes a cat a cat?

Hao Xu[a], Yuntian Chen [b, *], and Dongxiao Zhang[c,d,*]

[a] *BIC-ESAT, ERE, and SKLTCS, College of Engineering, Peking University, Beijing 100871, P. R. China*

[b] *Eastern Institute for Advanced Study, Yongriver Institute of Technology, Ningbo 315200, Zhenjiang, P. R. China*

[c] *National Center for Applied Mathematics Shenzhen (NCAMS), Southern University of Science and Technology, Shenzhen 518055, Guangdong, P. R. China*

[d] *Department of Mathematics and Theories, Peng Cheng Laboratory, Shenzhen 518000, Guangdong, P. R. China*

[*] Corresponding author.

E-mail address: 390260267@pku.edu.cn (H. Xu); ychen@eias.as.cn (Y. Chen); zhangdx@sustech.edu.cn (D. Zhang).

**Abstract:** The interpretability of deep neural networks has attracted increasing attention in recent years, and several methods have been created to interpret the "black box" model. Fundamental limitations remain, however, that impede the pace of understanding the networks, especially the extraction of understandable semantic space. In this work, we introduce the framework of semantic explainable AI (S-XAI), which utilizes row-centered principal component analysis to obtain the common traits from the best combination of superpixels discovered by a genetic algorithm, and extracts understandable semantic spaces on the basis of discovered semantically sensitive neurons and visualization techniques. Statistical interpretation of the semantic space is also provided, and the concept of semantic probability is proposed for the first time. Our experimental results demonstrate that S-XAI is effective in providing a semantic interpretation for the CNN, and offers broad usage, including trustworthiness assessment and semantic sample searching.

Convolutional neural networks (CNNs) have made tremendous progress in various fields, including computer vision[1], nature language processing[2], and other fields[3] in recent years. Although CNNs have achieved superior performance on varied tasks, the degree of confidence for the prediction is limited by its essence of a "black-box" model, which means that the decision process is difficult to express with explicit rules, and thus challenging to be understood by humans. This shortcoming will expose the CNN to the danger of being attacked or biased[4]. Therefore, the interpretability of the CNN has attracted increasing attention, and a variety of techniques have been attempted to explore the decision logic of CNNs in a human-understandable manner.

Interpretability is defined as the ability to provide explanations in understandable terms to a human[5]. Mainstream ways of interpreting CNNs include: (1) feature visualization, which visualizes specific filters or feature maps to depict the representation of a CNN[6–11]; (2) network diagnosis, which diagnoses a pre-trained CNN to understand different aspects of it[12–16]; and (3) structure modification, which adjusts the structure of CNN for better interpretability[17–21]. Despite their success, current interpretation methods face fundamental constraints that limit their usage. First, although some studies have focused on semantic information hidden behind the CNN[16,19,22], an explicit understandable semantic space has not yet been abstracted. Second, current researches are centered on extracting individual traits from each sample to obtain local interpretability, and are unable to decipher underlying common traits hidden in the data of the same class. Finally, even though existing works have provided numerous techniques to interpret CNN, few of them inspire applications or improvements in practical tasks. Overall, the study of the interpretation and application of semantic spaces in CNNs still needs to be further developed.

In order to overcome these existing limitations, this paper presents the semantic explainable AI (S-XAI), which is a semantic interpreter that provides a global interpretation by abstracting common traits from datasets and extracting explicit understandable semantic spaces from CNNs. It makes the following three notable improvements:

1. Global interpretation via common traits. Instead of seeking the information learned by CNNs in a single image, we adopt row-centered principal component analysis (PCA), rather than the commonly used column-centered PCA, to explore the common traits hidden behind samples that can be visualized by feature visualization techniques, which is illustrated in Fig. 1a.
2. Understandable semantic space. In this work, we extract understandable semantic spaces that can be visualized explicitly, which is shown in Fig. 1b. We also propose the concept of semantic probability for the first time, which can measure the probability of occurrence of semantic concepts.
3. Broad and efficient usage. The ultimate goal of understanding neural networks is to elucidate how they work and improve them. Our proposed S-XAI is able to handle more tasks efficiently, such as trustworthiness assessment and semantic sample searching, thus improving the confidence and scalability of CNNs, which is illustrated in Fig. 1c.

In this work, we take the task of discriminating between cats and dogs using the commonly used VGG-19 network as an example to demonstrate what makes a cat recognized as a cat for CNN from the perspective of semantic space. In the VGG-19 network, we add a global average pooling (GAP) layer before the fully-connected layer. The GAP layer reduces dimension, and preserves the spatial information extracted by the previous convolutional layers and pooling layers, which facilitates feature visualization[23]. In this paper, we first exhibit how to obtain underlying common traits, which are vectors containing mixed semantic features. Then, we describe how to extract understandable semantic space on the basis of the visualization of semantically sensitive neurons (SSNs) discovered from the comparison between the common traits of samples with masked and unmasked semantic concepts. Finally, we introduce the concept of semantic probability and discuss some major challenges for CNNs, including overconfidence for prediction[24] and target sample searching[25], and show how the proposed S-XAI can handle these issues in a simple yet effective way through several experiments.

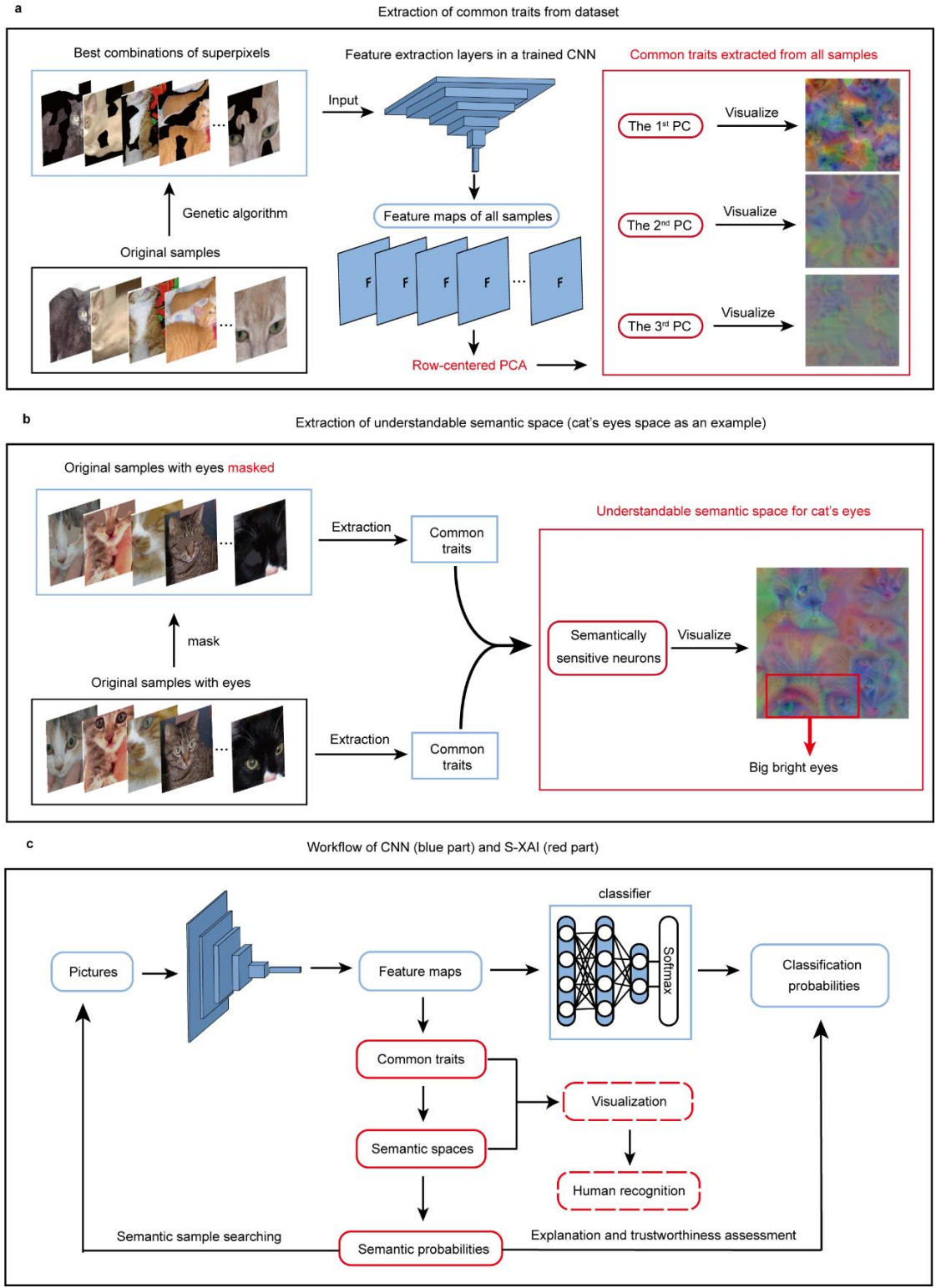

**Fig. 1.** Overview of the proposed S-XAI. **a,** Our framework for extracting common traits from datasets, taking the category of cats as an example. Left: the original samples and discovered best combinations of superpixels. Middle: extracting feature maps for all samples from a pre-trained CNN. Right: the obtained principal components (PCs) from the row-centered PCA on the feature maps and visualization of common traits. **b,** Our framework for extracting understandable semantic

space, taking the semantic space of cats' eyes as an example. Left: samples with unmasked and masked semantic concept. Middle: extraction of common traits for both kinds of samples. Right: discovered semantically sensitive neurons (SSNs) and the visualization of the semantic space. The big bright eyes are vividly illustrated, which proves that an understandable semantic space is found. **c,** The workflow of CNN and S-XAI. The blue part is the prediction process of the CNN. The red part is the process of S-XAI, in which the semantic probabilities are calculated from the extracted semantic spaces that can be visualized and recognized by humans for trustworthiness assessment and semantic sample searching. The dashed box refers to an optional step.

## Results
### Extraction of common traits from samples

In nature, the same kinds of objects often possess certain similar common characteristics, which are termed "common traits". These common traits constitute an important basis for the identification of species. For example, although there are numerous kinds of cats in the world, they all have similar faces, noses, and legs that are common traits for recognition as cats by humans. In previous literature, some techniques have been proposed to interpret CNNs by finding the part of an image with the largest response to the specified category[7,12]; however, this is a local interpretation since it only explains why one simple image is judged as a responsive label by the CNN. In other words, these studies have discovered individual traits, instead of common traits. In this work, we innovatively introduce the row-centered PCA to extract common traits from samples for CNNs. PCA is a widely used technique for reducing the dimensionality of datasets and increasing interpretability[26], which includes column-centered PCA and row-centered PCA. Compared with the commonly used column-centered PCA, row-centered PCA reorganizes the original samples into a new reduced sample space, which occurs relatively infrequently but is highly appropriate for extracting common traits in this work.

  The framework for extracting common traits from samples is illustrated in Fig. 1a. In the framework, $N_s$ different cat samples are randomly selected from the dataset, and a specific genetic algorithm is utilized to obtain the optimal combinations of superpixels for each sample, which aims to reduce interference and makes the extracted common traits more representative (see Supplementary Information). The discovered combinations of superpixels are fed into the CNN, and the feature maps are generated after the global average pooling (GAP) layer. Different from previous works that visualize feature maps for each sample[6,7], we conduct the row-centered PCA on the feature maps of all $N_s$ samples. Visualization of the 1st, 2nd, and 3rd PCs after row-centered PCA with 300 samples for cats and dogs is illustrated in Fig. 2a and 2b, respectively. It is obvious that different PCs present traits at different levels. The 1st PC displays vivid faces, the 2nd PC displays several body regions, such as beards, eyes and noses, and the 3rd PC mainly presents fur-like patterns. Different from the visualization of feature maps for one sample, the visualization of PCs contains multiple traits that clearly belong to different samples, which indicates that the PC integrates the common traits from the samples. The information ratios (calculated by the proportion of total variance) of the first five PCs are presented in Fig. 2c, which reveals that the 1st PC retains nearly 50% of the information, while the 2nd PC and 3rd PC retain only 7.32% and 3.53% of the information, respectively. This may explain why the 1st PC can display faces, which are a holistic concept, while the others can only present fragmental semantic parts. Considering the dominant information ratio of the 1st PC, the subsequent study will extract semantic spaces based on the 1st PC.

In order to further prove the stability of the extracted common traits, we conduct repeated experiments with different numbers of samples and calculate the spread from the average, which is expressed as:

$$e = \frac{1}{p}\frac{1}{N_e}\sum_{j=1}^{p}\sum_{i=1}^{N_e}\left|s_j^i - \bar{s}_j\right| \times 100\%,$$

where $N_e$ is the number of repeated experiments; $p$ is the number of features (i.e., the length of the PC); $s_j^i$ is the value of the $j^{th}$ index in the scores of PCs in the $i^{th}$ experiment; and $\bar{s}_j$ is the average value of the $j^{th}$ index in the scores of PC of all of the experiments, which is given as follows:

$$\bar{s}_j = \frac{1}{N_e}\sum_{i=1}^{N_e} s_j^i.$$

In this work, $N_e$ is 3 and $p$ is 512. The result is shown in Fig. 2d. From the figure, it is found that the spread tends to decrease as the number of selected samples $N_s$ increases, which indicates that the extracted traits tend to be more stable with more $N_s$. Specifically, when the $N_s$ is 300, the spread is only 3.9%, which means that we can extract stable common traits from a small number of samples, which further proves that certain representative common traits indeed exist in the samples of the same class.

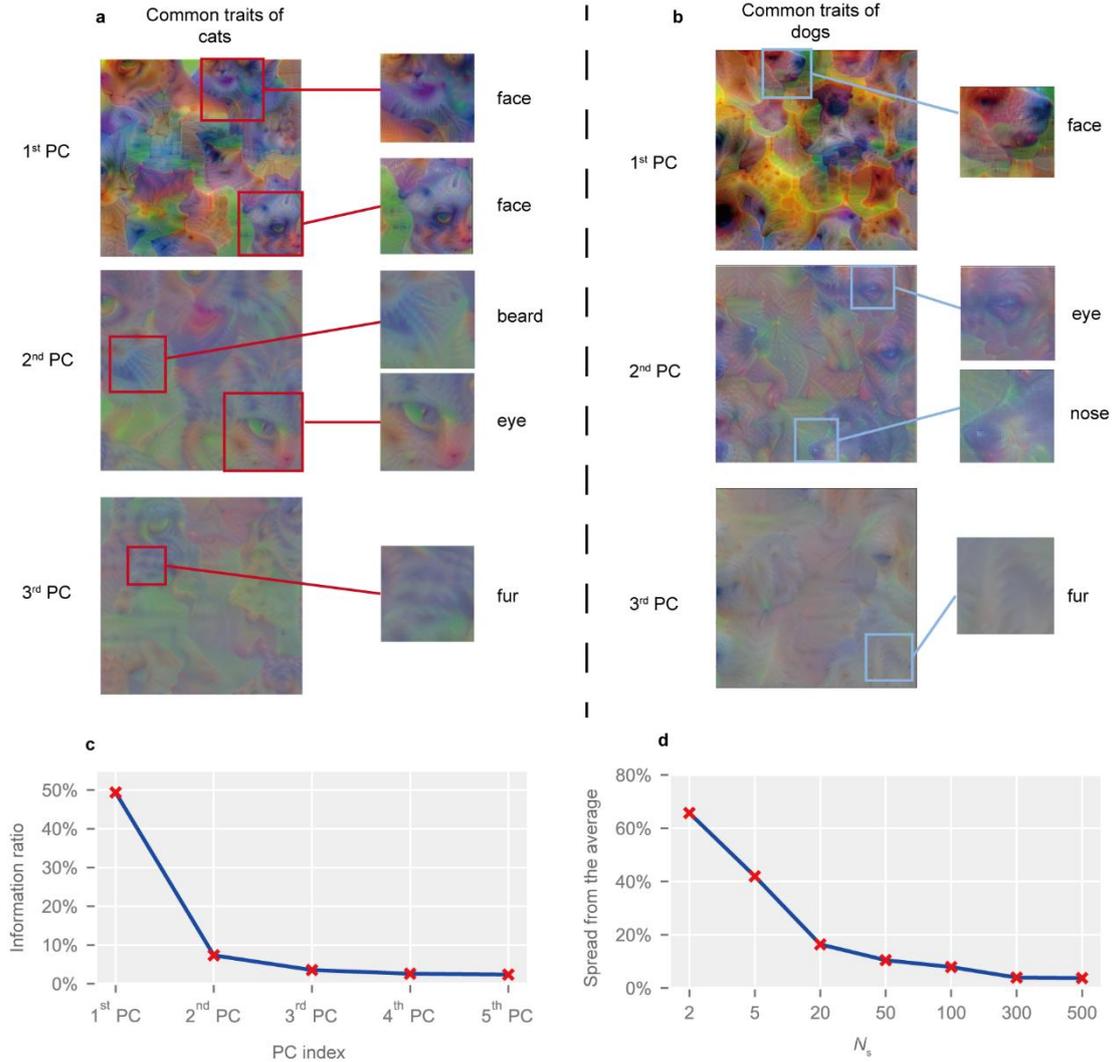

**Fig. 2. The results for extracting common traits. a,** Visualization of the 1st, 2nd, and 3rd PCs for cats ($N_s$=300). The left pictures are original visualized common traits, and the right ones are partial enlarged pictures with explicit traits for better recognition. **b,** Visualization of the 1st, 2nd, and 3rd PCs for dogs ($N_s$=300). The left pictures are original visualized common traits, and the right ones are partial enlarged pictures. **c,** The information ratio of the first five PCs extracted from 500 dog samples. The x-shape annotation refers to the information ratio. A higher information ratio means more feature information contained in the PC. **d,** The spread from the average for the 1st PC of cats with different $N_s$. A lower spread means a higher stability of extracted common traits.

**Extraction and visualization of understandable semantic space**

Although the common traits have been extracted successfully, the semantic information is disorganized since many kinds of semantic concepts are mixed in the extracted common traits. Therefore, we develop a framework to extract and visualize understandable semantic spaces on the basis of the common traits, which is illustrated in Fig. 1b. The core of our proposed framework is to compare the common traits extracted from samples with masked and unmasked semantic concepts. We take the semantic space of cats' eyes as an example. First, $N_s$ cat samples with visible eyes are chosen from the datasets, and their eyes are masked with surrounding color. Then, the respective common traits are extracted, as shown in Fig. 3a. A number of semantically sensitive neurons (SSNs), $N_{SSN}$, are chosen by selecting the maximum absolute difference between both common traits, which are visualized in Fig. 3a and 3b, respectively. The semantic concept is split successfully since there only exist multiple bright and clear eyes in the semantic space of cats' eyes, and the condition is the same for the semantic space of cats' nose. It is worth noting that the visualized eyes and noses are surreal, i.e., they are not obtained by visualizing a single sample, but rather by concretizing the entire semantic space. The results show that the representation of features at the semantic level from the CNN is successfully obtained.

In this work, the semantic concept is defined by humans to be better understood. However, in our research, we found that the connections between these semantic concepts seem to be more complex. For instance, the 263[th] neuron is sensitive to both cats' eyes and nose, which suggests that the network seems to learn the relationship between cats' eyes and nose. This is also proven by the fact that a blurry nose also appears in the semantic space of cats' eyes. This means that the semantic concepts learned by the network may be somewhat different from the semantic concepts recognized by humans, which calls for deeper investigation in future work.

**Statistical interpretation for semantic space**

In order to explore semantic space more deeply, we also provide statistical interpretation for semantic space. For the purpose of depicting the activation extent for an image $z$ in the semantic space $s$, we define a weighted average activation $A_s(z)$ as:

$$A_s(z) = \frac{1}{N_{SSN}} \sum_{i=1}^{N_{SSN}} a_i \Delta_i,$$

where $N_{SSN}$ is the number of discovered SSNs; $a_i$ is the activation of the $i$[th] SSN in the feature map extracted from the GAP layer before fully-connected layers for image $z$; and $\Delta_i$ is the value of difference for the $i$[th] SSN obtained from semantic space. A larger $A_s$ means that the image has a larger activation in semantic space $s$. Since the calculation process is very fast, the weighted average activation of a large number of samples can be quickly obtained. In this work, we randomly selected 3,000 samples containing a wide variety of cats from the dataset and calculated their weighted average activation in the extracted semantic space of cats' eyes. The probability density distribution plot of the values of the weighted average activation $A_s$ is displayed in Fig. 3c. The distribution is close to a normal distribution, which also occurs in other semantic spaces (see Supplementary Information). Furthermore, we find that the two ends of the probability density distribution correspond to cats without and with obvious eyes, respectively (see Fig. 3c). While images of cats without visible eyes are still able to be identified as cats with a high probability by the CNN, their activations within the semantic space of cats' eyes are low, which further confirms the validity of the extracted semantic space.

In order to measure the relative position of image $z$ in the distribution of samples in semantic space $s$, we propose the concept of semantic probability $P_s(z)$, which is written as follows:

$$P_s(z) = \frac{cdf(X = A_s(z)) - cdf(X = A_{min})}{cdf(X = A_{max}) - cdf(X = A_{min})}, \qquad (2)$$

where $cdf$ is the cumulative distribution function of the fitted normal distribution; and $A_{min}$ and $A_{max}$ are the min and max of $A_s$ in the distribution, respectively. A larger $P_s(z)$ represents a greater activation of image $z$ in the corresponding semantic space $s$ with a higher probability of occurrence.

From the figure, the $P_{eye}(z)$ of the picture with masked eyes exhibits a marked decrease compared with the same one with unmasked eyes in the semantic space of cats' eyes. Similarly, the images satisfying $P_{eye}(z)>0.9$ have large bright eyes, while the images satisfying $P_{eye}(z)<0.1$ hardly show any eyes.

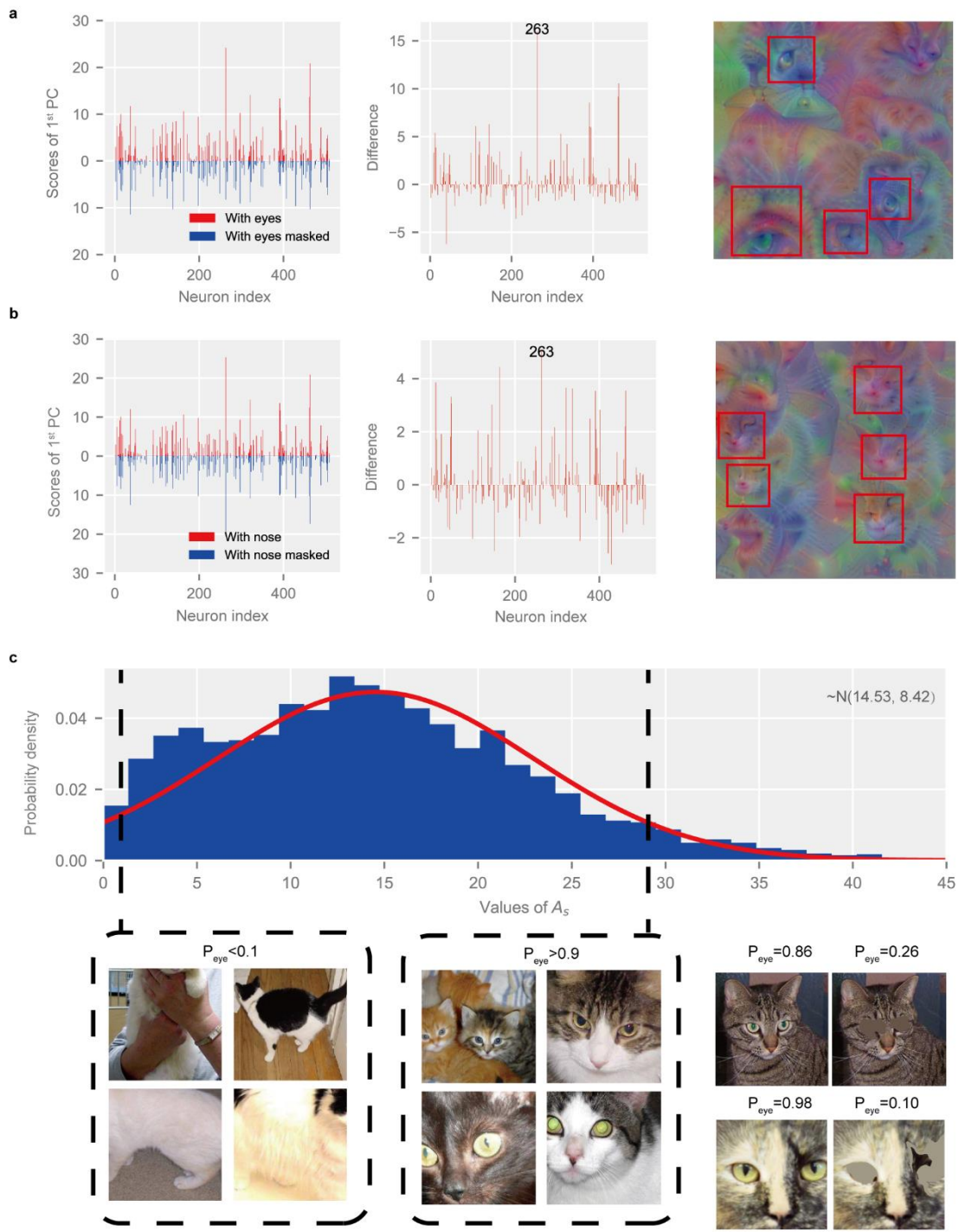

**Fig. 3. Results of extracted understandable semantic space. a,** Left: scores of the 1st PC (i.e., common traits) of samples with eyes unmasked and masked. Middle: difference between the extracted common traits. Right: visualization of extracted semantic space of cats' eyes, and the red frames are recognized eyes. **b,** Left: scores of the 1st PC of samples with nose unmasked and masked. Middle: difference between the extracted common traits. Right: visualization of extracted semantic

space of cats' nose, and the red frames are recognized noses. The annotation "263" in both **a** and **b** indicates that the same SSN occurs in both semantic spaces. **c,** Probability density distribution plot of the values of the weighted average activation $A_s$ for 3,000 samples of cats. The red curve is the fitted normal distribution curve. The pictures in the black framed lines are the samples located at the left ($P_{eye}$<0.1) and right ($P_{eye}$>0.9) ends of the distribution, respectively. The bottom right: the semantic probability of samples with eyes unmasked and masked in the semantic space of cats' eyes.

**Application: trustworthiness assessment**

In CNNs, the output of prediction is usually converted into the probability of each label by the softmax function, which will induce the problem of overconfidence since the softmax function exaggerates confidence[24]. For example, for a photograph of a cat's back shown in Fig. 4a, a human may experience a challenge distinguishing it from a blanket, but the CNN recognizes it as a cat with a high probability of 97.1%. However, the semantic space extracted in this work can solve this problem since the semantic space and respective semantic probability are obtained from the information in the feature maps, without needing to go through the fully-connected layers and softmax function, which means that the semantic probability is closer to the reality.

In this work, we extract six semantic spaces, including eyes, noses, and legs for cats and dogs, respectively. The semantic probabilities of these semantic spaces are illustrated by a radar map for each image. An explanation sentence is generated automatically by S-XAI from the radar map utilizing different words to represent different levels of confidence (e.g., sure, probably, may, confusing, etc.). For example, a Golden Retriever dog that is identified to be a dog with 100% probability by CNN is shown in Fig. 4b. The S-XAI can provide more information besides the classification probability given by the CNN. From the radar map, it is obvious that both the semantic probabilities of cats' legs and dogs' legs are high, which indicates that it is hard for the CNN to distinguish the legs of cats and dogs. In fact, if this dog's upper body is covered, humans cannot discern whether it is a dog or a cat by relying only on the legs. In addition, the semantic probabilities of dogs' eyes and noses are dominant compared with cats' eyes and noses. This information is reflected in the output of S-XAI that "*It is probably a dog mainly because its vivid eyes, which are something like dog's eyes. It has nose, which is something like dog's nose. However, its legs are a little confusing.*"

Another experiment is conducted on an image of a Maine Coon cat which is reported to be often mistaken for a dog due to its relatively large size. We select three pictures of this cat with different postures, including the front angle and the side angle (Fig. 4c), and the back angle (Fig. 4a). The output of CNN is similar and above 90% for the cats with all postures. However, radar maps obtained from the S-XAI identify differences between these pictures. For the image in the front angle, the explanation of S-XAI is "*I am sure it is a cat mainly because its eyes, and nose and legs, which are cat's eyes and nose and legs obviously*", which shows high confidence. For the image in the side angle, the explanation of S-XAI is "*It is probably a cat mainly because it has nose, which is perhaps cat's nose. Although its legs are a little confusing.*" For the image in the back angle, all semantic probabilities are inconspicuous, and the explanation of S-XAI is "*It may be a cat, but I am not sure.*" These explanations made by S-XAI are consistent with human cognition, in which from the front angle, it is universally identified to be a cat; whereas, its legs are confusing from the side angle, which may account for why this type of cat (Maine Coon) is frequently mistaken for a dog.

We also provide an example in which a white blanket is input into the CNN, which is shown in Extended data Fig. 1. While humans easily identify it as a white blanket instead of a cat, the CNN predicts that it has a 92.1% probability of being a cat, which is an incorrect assessment. The S-XAI, however, discovers that all semantic probabilities of this image are low and outputs that "*It may be a cat, but I am not sure*", which shows low confidence of the assessment. In contrast, when presented with a cat, S-XAI can provide an explanation with high confidence.

The above experiments show that the explanation of S-XAI provides more information to remedy the phenomenon of overconfidence and makes the process of prediction understandable for humans.

**Application: semantic sample searching**

Since current research on semantic interpretation of neural networks is still in its infancy, finding target samples by semantics requires additional effort[25]. However, once the semantic space is extracted, samples that satisfy semantic requirements through semantic probability can easily be found. For instance, if we want to find dogs with distinct noses in the dataset, we just need to set $P_{nose}^{dog}(z) > 0.9$ and the discovered samples all have obvious noses, which is illustrated in Fig. 4d. Similarly, images of cats without distinct legs can also be searched quickly by setting $P_{leg}^{cat}(z) < 0.1$. The searching process is very fast because semantic probability in the semantic space can be calculated simultaneously during the prediction process of the neural network. This technique shows promising potential in the filtering of datasets and identifying "bad samples".

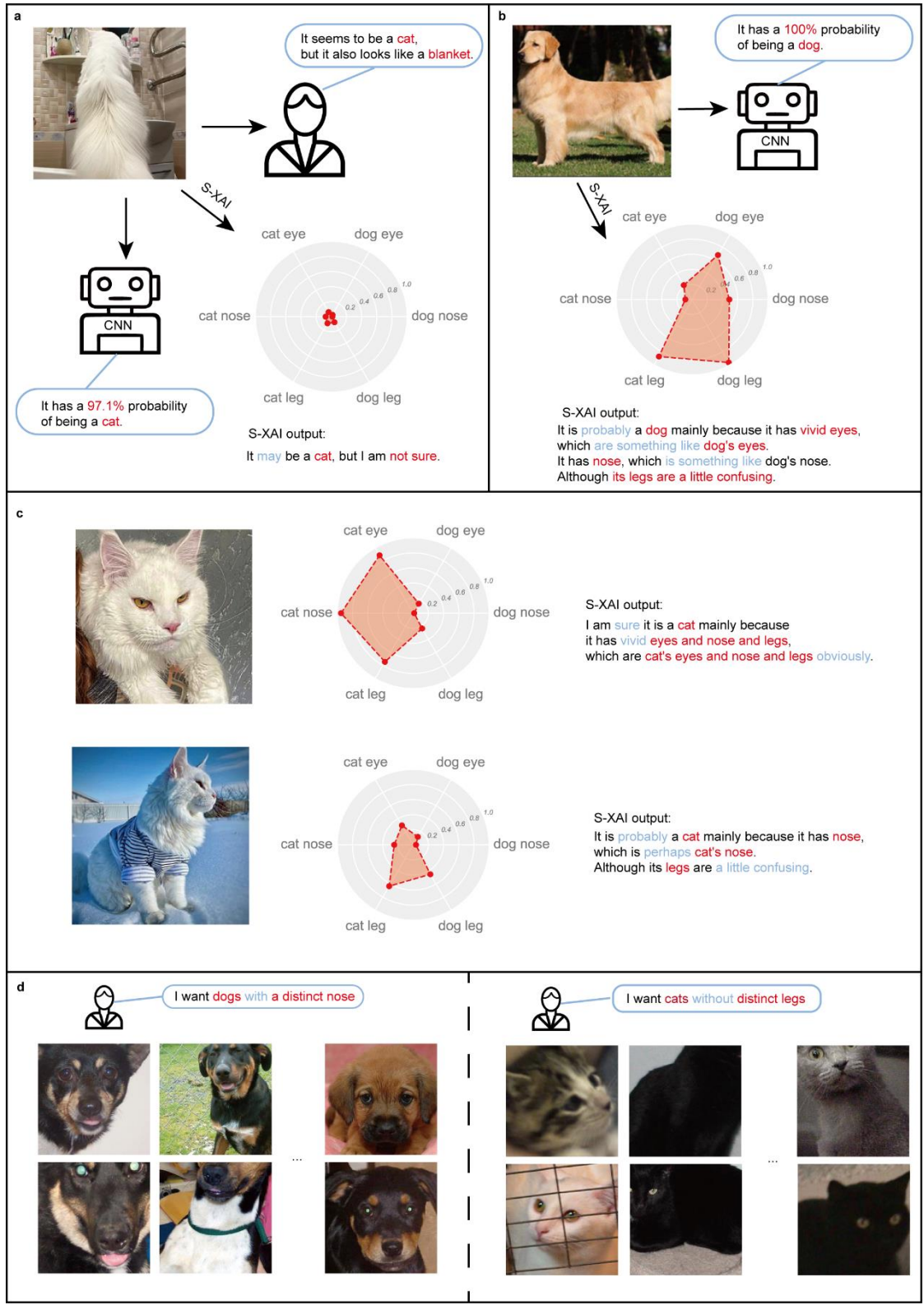

**Fig. 4. The application of S-XAI, including trustworthiness assessment and semantic sample searching. a,** Assessments given by humans (right), CNN (bottom), and S-XAI (bottom right) when identifying the picture of a Maine Coon cat facing backwards. The assessment of S-XAI includes a radar map of semantic probabilities in different semantic spaces and an explanation sentence

generated automatically from the semantic probabilities. **b,** Assessments given by CNN (right) and S-XAI (bottom) when identifying the picture of a Golden Retriever dog. **c,** Assessments given by S-XAI for the same Maine Coon cat with different postures and angles, including the front angle (upper) and the side angle (lower). **d,** The samples found by S-XAI that satisfy the semantic condition proposed by humans. Left: dogs with a distinct nose. Right: cats without distinct legs.

**Discussion**

In this work, we proposed the framework of the semantic explainable AI (S-XAI), which provides a global interpretation for CNNs by abstracting common traits from samples and extracting explicit understandable semantic spaces from CNNs. Statistical interpretation for the semantic space is further provided, and the concept of semantic probability is proposed for the first time. Experiments demonstrated that S-XAI is effective in providing a semantic interpretation for the CNN, and has broad usage, including trustworthiness assessment and semantic sample searching.

In S-XAI, the row-centered PCA plays a vital role, and can quickly extract highly hierarchical common traits from the feature maps of samples. Essentially, the row-centered PCA reorganizes the original samples into a new reduced sample space, where each PC corresponds to a certain level of feature information. Then, the semantic concept is separated from the common traits through discovering the semantically sensitive neurons (SSNs) and their specific proportional relationship. On this basis, we visualize the semantic space for the first time. A set of orthogonal semantic concepts, including eyes, nose and legs, is investigated in this work. It is found that the phenomenon of overlap of semantic spaces for CNN exists, which means that the semantic concepts in the CNN may be somewhat different from the definition of humans.

We also provide a statistical analysis for the extracted semantic space. The weighted average activation is defined in order to describe the activation of an image in semantic space, and it is discovered that the weighted average activation of sufficient natural samples is close to a normal distribution in the semantic space. Therefore, the concept of semantic probability is proposed for the first time, which can measure the likelihood of the occurrence of semantic concepts.

Our work goes a step further and investigates the application of our proposed S-XAI by several experiments. The results show that the S-XAI can provide trustworthiness assessment on the basis of semantic probabilities, which explores more information from the CNN, thus not only explaining the prediction made by the CNN, but also remedying the phenomenon of overconfidence. In addition, the S-XAI is proven to be able to quickly search for samples that satisfy semantic conditions. The extracted semantic space also sheds light on the identification of adversarial examples, which may provide a potential way for adversarial example defense (see Supplementary Information). It is also proven that the proposed S-XAI has good extensity when adapting it to other structures of CNN, such as AlexNet (see Supplementary Information).

Overall, our study enables exploration of the representation of CNNs at the semantic level, and provides an efficient way to extract understandable semantic space. The application of the S-XAI framework to other deep learning areas is relatively straightforward, and is currently being explored. Further work also involves simplifying the process of extracting semantic space by adopting certain techniques, such as semantic segmentation[27,28] or annotation-free techniques, for extracting object parts[16] to mask semantic concepts automatically and making semantic concepts more explicit to obtain better semantic space. We believe that the S-XAI will pave the way for understanding the "black box" model more deeply from the perspective of semantic space.

**Method**

**Superpixel segmentation.** Superpixel segmentation algorithms group pixels into perceptually meaningful atomic regions, which can be used to replace the rigid structure of the pixel grid[29]. This greatly reduces the complexity of image processing tasks. Here, the simple linear iterative clustering (SLIC) technique[29] based on k-means clustering is adopted for superpixel segmentation, the detailed process of which is provided in the Supplementary Information. Compared with most superpixel methods, SLIC has faster calculation speed, lower computational complexity, better stability, and is easy to implement in the python package *skimage*.

**Genetic algorithm.** The genetic algorithm is a well-known optimization algorithm inspired from the biological evolution process[30]. A typical genetic algorithm consists of crossover, mutation, fitness calculation, and evolution[31]. In this work, a specific genetic algorithm is put forward to find the best combinations of superpixels based on the segmentation of superpixels that incurs the highest probability of the specified category from the CNN. For an image $z$, we split it into $N_{sp}$ superpixels through the SLIC method and number each superpixel. A binary code of length $N_{sp}$ is generated as the genome according to the superpixels, where 1 represents the existence of the superpixel and 0 represents non-existence. For the initial generation, $N_P$ genomes are randomly initialized. Then, the crossover process is conducted by randomly exchanging certain segments of two genomes, which leads to the occurrence of new genomes. Afterwards, the genomes are mutated by randomly selecting genes and conducting bitwise negation. The fitness is then calculated to distinguish between good and bad genomes, which is defined as $P_{i=c}(z')$, where $z'$ is the combination of superpixels translated from the genome, $c$ is the target class, and $P_{i=c}$ is the output probability of the class $c$ obtained from the CNN. Next, the genomes are sorted from large to small according to the calculated fitness. Finally, the first half of genomes remain, while others are replaced by new genomes in order to produce a new generation of parents. This process will cycle until the epoch achieves the maximum iteration that is set beforehand. In order to accelerate the convergence, the best genome is fixed until a better one replaces it in each generation. In this work, $N_{sp}$ is 40, the population of genomes $N_P$ is 50, the maximum iteration is 50, and the probability of mutation is 0.5.

**Dogs vs. Cats dataset.** We used a well-known dataset originated from the competition in Kaggle called Dogs v.s. Cats. This dataset includes 25,000 training images in which half are cats and half are dogs, and 12,500 test images. We rescaled all of the images into normalized 224×224×3 pixels to use the VGG-19 network[31] as a classifier. In the VGG-19 network, we add a global average pooling (GAP) layer before the fully-connected layer. We used the ImageNet-pre-trained weights as initial weights, as well as the Adam optimizer[32] to train the neural network for three epochs through the mini-batch technique with a batch size of 16. The accuracy of the network on the test dataset is 97.5%.

**Row-centered PCA.** The PCA method involves a dataset with observations on $p$ numerical variables or features, for each of $n$ individuals or observations, which forms an $n \times p$ data matrix $X$[26].

In this work, $n$ equals the number of samples $N_s$, and $p$ is the number of features which equals the number of channels in this work. For a standard PCA that is usually utilized to reduce dimension, the columns of the data matrix are centered to calculate the covariance matrix, which facilitates dimension reduction after singular value decomposition (SVD). Therefore, the standard PCA is also termed as the column-centered PCA. In this work, we use the row-centered PCA, which is relatively less used, but achieves good performance on extracting a reduced sample space. In the row-centered PCA, the rows of the data matrix $X$ are centered. We denote the row-centered data matrix $\hat{X}$ where $\hat{X}_{i,j} = X_{i,j} - \overline{X}_i$, $i$ is the index of rows, and $j$ is the index of columns. The covariance matrix $S$ is calculated as $S = \frac{1}{p-1}\hat{X}\hat{X}^T$. Then, the SVD is adopted to the covariance matrix $S$, and $S = U\Sigma V^T$ is obtained where $U, \Sigma, V \in R^{n \times n}$. $\Sigma$ is a diagonal matrix, called the singular value matrix, where the diagonal elements (i.e., singular value) are arranged from largest to smallest. Since $\Sigma$ is a square matrix, the singular value equals the eigenvalue. It is worth noting that the rank of $\Sigma$ is $r < \min\{n, p\}$, which means that there are only $r$ non-zero singular values (or eigenvalues). Herein, we preserve the first $k$ ($k \leq r$) singular values and obtain $U_k \in R^{n \times k}$ from $U$ by retaining the first $k$ columns. Finally, the reduced principal components (PCs) are obtained by $X_k = X^T U_k \in R^{p \times k}$ where each column of $X_k$ is a PC. For example, $X_k^{(col=1)}$ is the 1st PC, $X_k^{(col=i)}$ is the $i$th PC, and the elements of PC are called PC scores. The standard measure of quality of a given $i$th PC is the proportion of total variance (or the information ratio) that is calculated as $\pi_i = \lambda_i \left( \sum_{i=1}^{p} \lambda_i \right)^{-1} = \lambda_i (\text{tr}(S))^{-1} \times 100\%$, where tr($S$) denotes the trace of the covariance matrix S, and $\lambda_i$ is the corresponding eigenvalue. Therefore, the quality of the retained $k$-dimensional PCs can be expressed as a percentage of total variance accounted for: $\sum_{i=1}^{k} \pi_i$. It is common practice to use some predefined percentage of total variance explained to decide how many PCs should be retained (70% of total variability is common), and the emphasis in PCA is almost always on the first few PCs[26].

**Feature visualization.** The feature visualization techniques aim to solve the following question: given an encoding feature of an image, to what extent is it possible to reconstruct the image itself?[6] This question is closely related to the explanation of the network since the visualized feature map can tell us what kinds of features are learned by the network. The core issue of feature visualization is solving the minimization problem: $z^* = \arg\min_{z \in R^{C \times H \times W}} \left( L(\Phi(z), \Phi_0) + \lambda R(z) \right)$, where $L$ is the loss function which is usually mean squared error (MSE), $\Phi_0$ is the target encoding feature, and $\Phi(z)$ is the corresponding feature of optimized image $z$ with the size of $C\times H\times W$ obtained from the network, where $C$ is the number of channels, $H$ and $W$ are the height and width of the feature map, respectively, and $R$ is the regularization term capturing a natural image prior. The focal point of the minimization is the selection of $R$, and thus leads to different techniques of feature visualization, including L$_2$-regularization[6], total variation (TV) regularization[6], and deep image prior regularization[10]. The existence of $R$ prevents the minimization process from converging to the images of high frequency that humans cannot discern. In this work, we utilize MSE loss for $L$ and total variation (TV) regularization for $R$. Therefore, the minimization problem is converted into the following:

$$z^* = \arg\min_{z \in R^{C\times H\times W}} \left( \|\Phi(z) - \Phi_0\|^2 + \lambda R(z) \right),$$

$$R(z) = \sum_k \sum_{i,j} \left( (z_{k,i,j+1} - z_{k,i,j})^2 + (z_{k,i+1,j} - z_{k,i,j})^2 \right)^{\frac{\beta}{2}}.$$

where $\lambda$ and $\beta$ are employed to control the magnitude of regularization. The gradient descent technique is used to solve this minimization problem as follows:

$$z^{n+1} = z^n - \delta \cdot \frac{d(\|\Phi(z^n) - \Phi_0\|^2 + \lambda R(z^n))}{dz^n},$$

where $z^n$ is the image after the $n^{th}$ iteration; $\delta$ is the learning rate; and the initial image is defined as a blank figure with all elements equal to zero after standardization. The optimization process will continue until the epoch reaches the maximum iteration. In this work, $\lambda$, $\beta$, and $\delta$ are set to be 2, 2, and 0.05, respectively. The learning rate $\delta$ is multiplied by 0.5 for each 1,000 epochs. The maximum iteration is 4,000.

In previous works[6,10,11], feature visualization techniques are usually employed to visualize the feature maps in the network, which means that they can only analyze what is learned by the network for each image. Although they enable understanding the feature learned by the network in each layer, they are also constrained to be a local explanation. In this work, we do not focus on the visualization of each single image, but rather visualization of the principal components obtained by the row-centered PCA that contains common traits, thus providing a global interpretation. Therefore, the $\Phi_0$ is each PC and the optimized image $z$ is the visualization of the PC, which contains common traits in this work.

**Extraction of common traits.** With the assistance of the row-centered PCA, it is possible to extract common traits from the samples of the same class. We extract common traits for cats as an example (see Fig. 1a). We first randomly select $N_s$ cat samples from the dataset, and discover the best combination of superpixels for each sample via the SLIC technique. Subsequently, the samples are fed into the network, and the feature maps of the last layer before the fully-connected classifier are extracted. The size of the feature map for each sample is a $C\times H\times W$ matrix. Specifically, the feature map is degenerated to a vector with the length of $C$ in this work since the last layer is the global average pooling (GAP) layer. Consequently, the feature maps for the selected $N_s$ cat samples form a matrix with the size of $N_s \times C$. In order to extract common traits, the row-centered PCA is adopted. Considering that the number of channels $C$ equals the number of features $p$ in this work, the matrix of feature maps is seen as the data matrix with the size of $N_s \times p$. In the row-centered PCA, we preserve $k$ PCs that make the percentage of total variance achieve 85% of the total variability and obtain a $k \times p$ matrix after PCA. In this work, we primarily focus on the first few PCs, since the information ratios of others are negligible. This row-centered PCA technique can also be utilized for other layers or networks easily (see Supplementary Information). After the row-centered PCA, the feature visualization technique is adopted to visualize each PC to exhibit the common traits in a human-understandable way (see the right of Fig. 1a).

**Masking the semantic concept.** In order to explore the semantic space, it is necessary to mask the semantic concept, such as eyes, nose, and legs in the image (see the left of Fig. 1b). In this work, it is completed manually since the semantic concept is defined by humans. To improve the efficiency of the masking process, we design a program to assist manual processing. For an image $z$, it is firstly split into $N_{sp}$ superpixels ($N_{sp}$=20 here). Then, the superpixels that contain the target semantic concept are selected manually, and are filled with the color of a nearby superpixel that is also selected manually. This process will mask the semantic concept, thus obtaining the same image without the target semantic concept. In this work, we have shown that 100 images with and without the target semantic concept are sufficient for extracting the semantic space, which means that manual processing is acceptable here. However, when faced with extraction of a large-scale semantic space with numerous semantic concepts and categories, manual processing is inefficient and some other techniques, such as semantic segmentation[27,28] or annotation-free techniques, for extracting object parts[16] may be viable approaches to improve this process.

**Extraction and visualization of semantic space.** In this work, for the first time, a simple yet effective method to extract and visualize semantic space in an understandable manner is presented. Here, we will introduce the method in detail. Our method is constructed based on the extracted common traits of the images with and without the target semantic concept (see Fig. 1b). First, $N_s$ samples with the explicit target semantic concept are selected, and then the target semantic concept is masked to generate samples without the target semantic concept ($N_s$=100 in this work). For masked and unmasked samples, the common traits are extracted in the same way as that mentioned above, respectively. It is worth noting that we do not need to search the best combinations of superpixels here since we focus on the semantic concept. Considering that the information ratio of the 1st PC is dominant compared with that of the others, which means that it contains more information of common traits, we only analyze the 1st PC here. The difference between both common traits is calculated as $s_{unmask}$-$s_{mask}$, where $s_{mask}$ and $s_{unmask}$ are the scores of the 1st PC for the samples with masked and unmasked semantic concept, respectively. From the comparison, it is found that several neurons are particularly sensitive to the existence of the semantic concept, which are termed as semantically sensitive neurons (SSNs) in this work. $N_{SSN}$ semantically sensitive neurons with the largest absolute difference are chosen. It is worth mentioning that the absolute difference of these SSNs constitutes a specific proportional relationship. The extracted semantic space is made up of both the SSNs and the specific proportional relationship. In this work, $N_{SSN}$ is set to be 5. In order to visualize the extracted semantic space, we design a target $\Phi_0$ for visualization, where its size is the same as the feature map in the last layer before fully-connected layers, and all elements are zeros except the discovered SSNs. For the discovered SSNs in $\Phi_0$, the values obey the specific proportional relationship with the maximum value enlarged (to 30 in this work). Through the visualization of $\Phi_0$, the semantic space is exhibited in a human-understandable way (see the right of Fig. 1b).

**Trustworthiness assessment.** The proposed S-XAI is able to provide more information and make the trustworthiness assessment based on the extracted semantic space. The trustworthiness assessment includes two components: a radar map of semantic probabilities, and a trustworthiness assessment. The radar map depicts the semantic probabilities in different semantic spaces. Through the radar map, a substantial amount of information can be observed to produce the trustworthiness assessment. Here, for an image *z*, we first focus on the semantic space of the predicted class given by the CNN, in which the maximum semantic probability is denoted as $P_{max}(z)$ and the respective semantic concept is denoted as $S_{max}$. Then, for each semantic concept, the class that has the maximum semantic probability (except the predicted class) is discovered, and the difference of the semantic probabilities between the predicted class and the discovered class are defined as $\Delta P(z)$. Among $\Delta P(z)$ for all semantic spaces, the maximum is termed as $\Delta_{max}P(z)$. The standard of a trustworthiness assessment is determined by $P_{max}(z)$, $\Delta P(z)$ and $\Delta_{max}P(z)$, which identify the confidence by using different words (e.g., sure, probably, may, confusing, etc.). The rules for generating the explanation by S-XAI is provided in Extended Data Table 1. It is worth noting that the explanation given by S-XAI is generated automatically without any additional information or manual processing. Once the semantic spaces have been extracted and the distribution of the weighted average activation has been fixed, our proposed S-XAI can make the trustworthiness assessment and provide a corresponding explanation as soon as the network makes the prediction.

**Data availability**
The user study data are available at Google Drive Share: https://drive.google.com/file/d/1OnEX67h-C7Q0_Tul_3TYousoTRW_Xd69/view?usp=sharing.

**Code availability**
The code to reproduce the figures and results of this article can be found at https://github.com/woshixuhao/semantic-explainable-AI. The pre-trained models and some results are provided by Google Drive Share at:
https://drive.google.com/file/d/1sXPUR3dwBE1HjykqO4ftwpAIHCqErY16/view?usp=sharing
https://drive.google.com/file/d/1Eythcx_wj-_eZ9JIaQtKwHyITzEKwtQ5/view?usp=sharing

**Contributions**
HX, YC, and DZ conceived the project, designed and performed research, and wrote the paper; HX implemented workflow, created code, visualized results, analysed and curated data.

**Competing interests**
The authors declare no competing interests.

**References**
1. Krizhevsky, A., Sutskever, I. & Hinton, G. E. ImageNet classification with deep convolutional

neural networks. *Commun. ACM* **60**, (2017).

2. Gehring, J., Auli, M., Grangier, D., Yarats, D. & Dauphin, Y. N. Convolutional sequence to sequence learning. in *34th International Conference on Machine Learning, ICML 2017* vol. 3 (2017).

3. Gu, J. *et al.* Recent advances in convolutional neural networks. *Pattern Recognit.* **77**, 354–377 (2018).

4. Szegedy, C. *et al.* Intriguing properties of neural networks. in *2nd International Conference on Learning Representations, ICLR 2014 - Conference Track Proceedings* 1–10 (2014).

5. Zhang, Y., Tino, P., Leonardis, A. & Tang, K. A Survey on Neural Network Interpretability. *IEEE Transactions on Emerging Topics in Computational Intelligence* vol. 5 726–742 (2021).

6. Mahendran, A. & Vedaldi, A. Understanding deep image representations by inverting them. in *Proceedings of the IEEE Computer Society Conference on Computer Vision and Pattern Recognition* vols. 07-12-June-2015 (2015).

7. Selvaraju, R. R. *et al.* Grad-CAM: Visual Explanations from Deep Networks via Gradient-Based Localization. *Int. J. Comput. Vis.* **128**, 336–359 (2020).

8. Erhan, D., Bengio, Y., Courville, A. & Vincent, P. Visualizing higher-layer features of a deep network. *Bernoulli* 1–13 (2009).

9. Zhou, B., Khosla, A., Lapedriza, A., Oliva, A. & Torralba, A. Learning Deep Features for Discriminative Localization. in *Proceedings of the IEEE Computer Society Conference on Computer Vision and Pattern Recognition* vols 2016-Decem 2921–2929 (2016).

10. Lempitsky, V., Vedaldi, A. & Ulyanov, D. Deep Image Prior. in *Proceedings of the IEEE Computer Society Conference on Computer Vision and Pattern Recognition* 9446–9454 (2018). doi:10.1109/CVPR.2018.00984.

11. Dosovitskiy, A. & Brox, T. Inverting visual representations with convolutional networks. in *Proceedings of the IEEE Computer Society Conference on Computer Vision and Pattern Recognition* vols 2016-Decem 4829–4837 (2016).

12. Ribeiro, M. T., Singh, S. & Guestrin, C. 'why should i trust you?' explaining the predictions of any classifier. in *NAACL-HLT 2016 - 2016 Conference of the North American Chapter of the Association for Computational Linguistics: Human Language Technologies, Proceedings of the Demonstrations Session* 97–101 (2016). doi:10.18653/v1/n16-3020.

13. Bau, D., Zhou, B., Khosla, A., Oliva, A. & Torralba, A. Network dissection: Quantifying interpretability of deep visual representations. in *Proceedings - 30th IEEE Conference on Computer Vision and Pattern Recognition, CVPR 2017* vols 2017-January (2017).

14. Lu, Y. Unsupervised learning on neural network outputs: With application in zero-shot learning. in *IJCAI International Joint Conference on Artificial Intelligence* vols 2016-Janua 3432–3438 (2016).

15. Aubry, M. & Russell, B. C. Understanding deep features with computer-generated imagery. in *Proceedings of the IEEE International Conference on Computer Vision* vol. 2015 Inter 2875–2883 (2015).

16. Zhang, Q., Wang, X., Cao, R. & Wu, Y. N. Extraction of an Explanatory Graph to Interpret a CNN. *IEEE Trans. Pattern Anal. Mach. Intell.* **43**, 3863–3877 (2021).

17. Wu, M. *et al.* Beyond sparsity: Tree regularization of deep models for interpretability. in *32nd AAAI Conference on Artificial Intelligence, AAAI 2018* 1670–1678 (2018).

18. Zhang, Q., Yang, Y., Ma, H. & Wu, Y. N. Interpreting cnns via decision trees. in *Proceedings*

*of the IEEE Computer Society Conference on Computer Vision and Pattern Recognition* vols 2019-June 6254–6263 (2019).

19. Zhang, Q., Wu, Y. N. & Zhu, S. C. Interpretable Convolutional Neural Networks. in *Proceedings of the IEEE Computer Society Conference on Computer Vision and Pattern Recognition* 8827–8836 (2018). doi:10.1109/CVPR.2018.00920.

20. Zhang, Q., Wang, X., Wu, Y. N., Zhou, H. & Zhu, S. C. Interpretable CNNs for Object Classification. *IEEE Trans. Pattern Anal. Mach. Intell.* **43**, 3416–3431 (2021).

21. Blazek, P. J. & Lin, M. M. Explainable neural networks that simulate reasoning. *Nat. Comput. Sci.* **1**, 607–618 (2021).

22. Vaughan, J., Sudjianto, A., Brahimi, E., Chen, J. & Nair, V. N. Explainable Neural Networks based on Additive Index Models. Preprint at http://export.arxiv.org/abs/1806.01933 (2018).

23. Lin, M., Chen, Q. & Yan, S. Network in network. *2nd Int. Conf. Learn. Represent. ICLR 2014 - Conf. Track Proc.* 1–10 (2014).

24. Guo, C., Pleiss, G., Sun, Y. & Weinberger, K. Q. On calibration of modern neural networks. *34th Int. Conf. Mach. Learn. ICML 2017* **3**, 2130–2143 (2017).

25. Allot, A. *et al.* LitVar: A semantic search engine for linking genomic variant data in PubMed and PMC. *Nucleic Acids Res.* **46**, W530–W536 (2018).

26. Jollife, I. T. & Cadima, J. Principal component analysis: A review and recent developments. *Philosophical Transactions of the Royal Society A: Mathematical, Physical and Engineering Sciences* vol. 374 (2016).

27. Girshick, R., Donahue, J., Darrell, T. & Malik, J. Rich feature hierarchies for accurate object detection and semantic segmentation. in *Proceedings of the IEEE Computer Society Conference on Computer Vision and Pattern Recognition* (2014). doi:10.1109/CVPR.2014.81.

28. Long, J., Shelhamer, E. & Darrell, T. Fully convolutional networks for semantic segmentation. in *Proceedings of the IEEE Computer Society Conference on Computer Vision and Pattern Recognition* vols 07-12-June-2015 (2015).

29. Achanta, R. *et al.* SLIC superpixels compared to state-of-the-art superpixel methods. *IEEE Trans. Pattern Anal. Mach. Intell.* **34**, 2274–2281 (2012).

30. Katoch, S., Chauhan, S. S. & Kumar, V. A review on genetic algorithm: past, present, and future. *Multimed. Tools Appl.* **80**, (2021).

31. Whitley, D. A genetic algorithm tutorial. *Stat. Comput.* **4**, 65–85 (1994).

32. Kingma, D. P. & Ba, J. L. Adam: A method for stochastic optimization. in *3rd International Conference on Learning Representations, ICLR 2015 - Conference Track Proceedings* (2015).

**Extended Data**

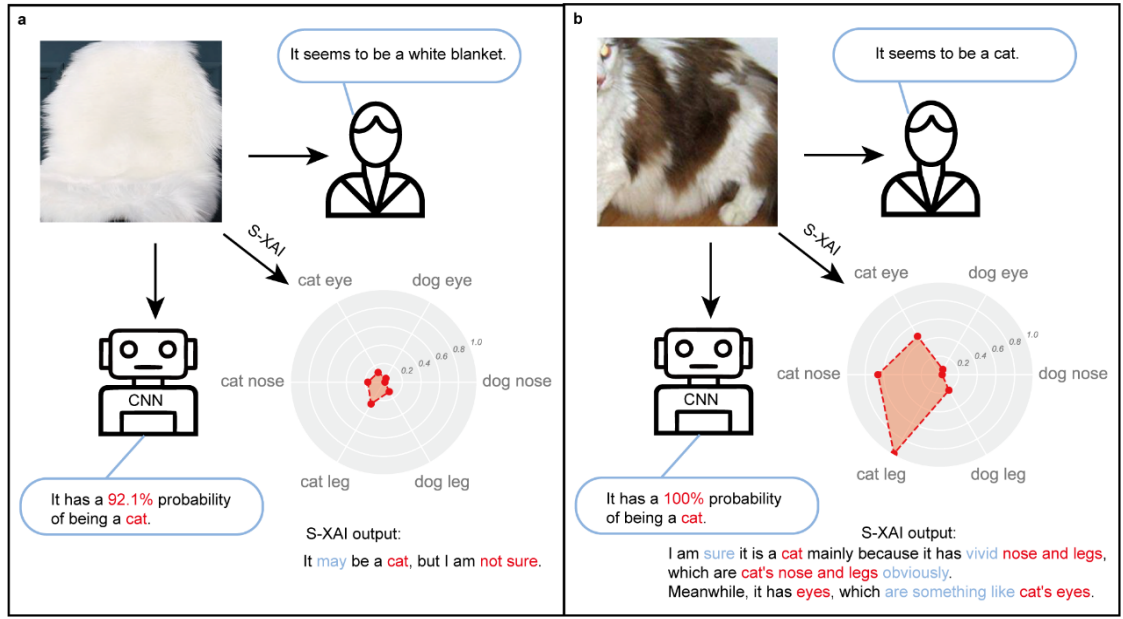

**Extended Data Fig. 1.** Assessments given by humans (right), CNN (bottom), and S-XAI (bottom right) when identifying the picture of a white blanket (a) and a cat (b).

**Extended Data Table 1.** Rules for generating the explanation by S-XAI.

|  | $\Delta_{max}P(z)<0.2$ | $0.2<\Delta_{max}P(z)<0.35$ | $0.35<\Delta_{max}P(z)<0.5$ | $\Delta_{max}P(z)>0.5$ |
|---|---|---|---|---|
| **Assessment** | It might be a dog/cat, but I am not sure. | I am sure it is a dog/cat mainly because | | It is probably a dog/cat mainly because |
| **Explanation** | | | | |
|  | $\Delta P(z)<0.2$ | $0.2<\Delta P(z)<0.35$ | $0.35<\Delta P(z)<0.5$ | $\Delta P(z)>0.5$ |
| $P_{max}(z)>0.5$ | Position: **vivid**, Semanteme: **confusing** | Position: **vivid**, Semanteme: **perhaps** | Position: **vivid**, Semanteme: **something like** | Position: **vivid**, Semanteme: **obviously** |
| $0.2<P_{max}(z)<0.5$ | **None** | Position: **be**, Semanteme: **perhaps** | Position: **be**, Semanteme: **something like** | Position: **be**, Semanteme: **obviously** |

# Supplementary Information for

# Semantic interpretation for convolutional neural networks:

# What makes a cat a cat?

Hao Xu[a], Yuntian Chen [b, *], and Dongxiao Zhang[c,d,*]

[a] *BIC-ESAT, ERE, and SKLTCS, College of Engineering, Peking University, Beijing 100871, P. R. China*

[b] *Eastern Institute for Advanced Study, Yongriver Institute of Technology, Ningbo 315200, Zhenjiang, P. R. China*

[c] *National Center for Applied Mathematics Shenzhen (NCAMS), Southern University of Science and Technology, Shenzhen 518055, Guangdong, P. R. China*

[d] *Department of Mathematics and Theories, Peng Cheng Laboratory, Shenzhen 518000, Guangdong, P. R. China*

[*] Corresponding author.

**The PDF file includes:**

**S.1**: Supplementary information for the extraction of common traits.

**S.2**: Extension of the row-centered PCA.

**S.3**: Supplementary information for the statistical explanation of semantic space.

**S.4**: Expansibility of the S-XAI to other structures of CNN.

**S.5**: Adversarial example identification.

**S.6:** Supplementary discussions

**S.7:** Details of superpixel segmentation

**Supplementary Fig. 1**. Comparison between the common traits extracted from samples with and without experiencing the genetic algorithm.

**Supplementary Fig. 2**. Visualization of the last PC in the row-centered PCA.

**Supplementary Fig. 3**. Scores of the 1$^{st}$ PC with different $N_s$ utilizing different random seeds to select different samples.

**Supplementary Fig. 4**. The information ratio of the 1$^{st}$ PC for each layer in the network.

**Supplementary Fig. 5**. Probability density distribution plots and q-q plots for different semantic

spaces.

**Supplementary Fig. 6**. The results of S-XAI for AlexNet.

**Supplementary Fig. 7**. Assessments given by humans, CNN, and S-XAI when identifying the true sample and the adversarial example.

**Supplementary Table 1**. The symbols used in this work and their meanings.

**Supplementary Table 2**. The success rate of attack by PGD and the success rate of defense by S-XAI with different attack strengths.

**Supplementary Table 1.** The symbols used in this work and their meanings.

| Symbol | Meaning |
| --- | --- |
| $N_s$ | The number of selected samples |
| $N_{SSN}$ | The number of semantically sensitive neurons |
| $A_s(z)$ | The weighted average activation of image $z$ |
| $P_s(z)$ | The semantic probability of image $z$ in semantic space $s$ |
| $N_e$ | The number of repeated experiments |
| $p$ | The number of features |
| $C$ | The number of channels |
| $H, W$ | The height and width of the feature map, respectively |
| $N_{sp}$ | The number of superpixels |
| $N_P$ | The initial population of genomes |
| $P_{i=c}(z)$ | The output probability of the class $c$ obtained from CNN |
| $e$ | Spread from the average |

**S.1**: **Supplementary information for the extraction of common traits**

In this work, a specific genetic algorithm is utilized to obtain the optimal combinations of superpixels for each sample. Here, an experiment is conducted to compare the common traits extracted from samples with and without experiencing the genetic algorithm, the results of which are displayed in Supplementary Fig. 1. From the figure, it is obvious that the common traits extracted from the best combinations of superpixels discovered by the genetic algorithm present more explicit semantic concepts compared with those without experiencing the genetic algorithm, which proves that the genetic algorithm assists to reduce interference and makes the extracted common traits more representative.

From the visualization of the 1st, 2nd, and 3rd PCs after the row-centered PCA, it is discovered that different PCs present traits at different levels. Considering that the first several PCs contain a large number of common traits, it is interesting to visualize the last PC, which is shown in Supplementary Fig. 2. Here, we retain 299 PCs from 300 samples and the last PC is the 299th PC. From the figure, high-heel shoes and a medicine bottle that constitute the background of the main images emerge in the visualization, which implies that the information ratio is closely related to the concentration of the common features.

Here, we also provide the scores of the 1st PC with different $N_s$ utilizing different random seeds to select different samples, and the results are shown in Supplementary Fig. 3. From the figure, it can be seen that the 1st PC is more stable when the $N_s$ is larger. Furthermore, it is discovered that the scores of the 1st PC exhibit a trend of proportional expansion and maintain a constant proportional relationship between the scores when $N_s$ increases. This implies that the constant proportional relationship determines the content of the common traits, while the magnitude of the scores determines the number of common traits presented by the visualization.

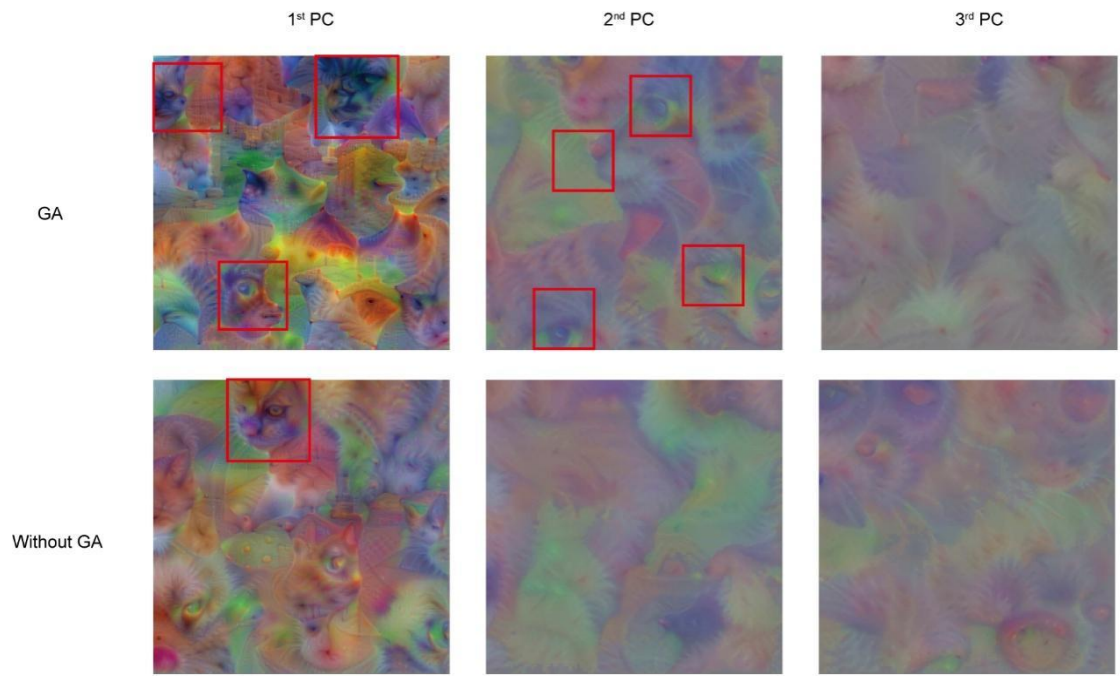

**Supplementary Fig. 1**. Comparison between the common traits, including the 1st PC, 2nd PC, and 3rd PC extracted from samples with and without experiencing the genetic algorithm. The red frame refers to the semantic concepts that can be recognized explicitly.

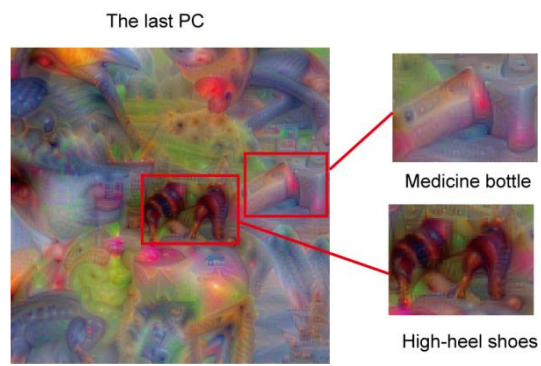

**Supplementary Fig. 2**. Visualization of the last PC (left) in the row-centered PCA and partial enlarged pictures (right).

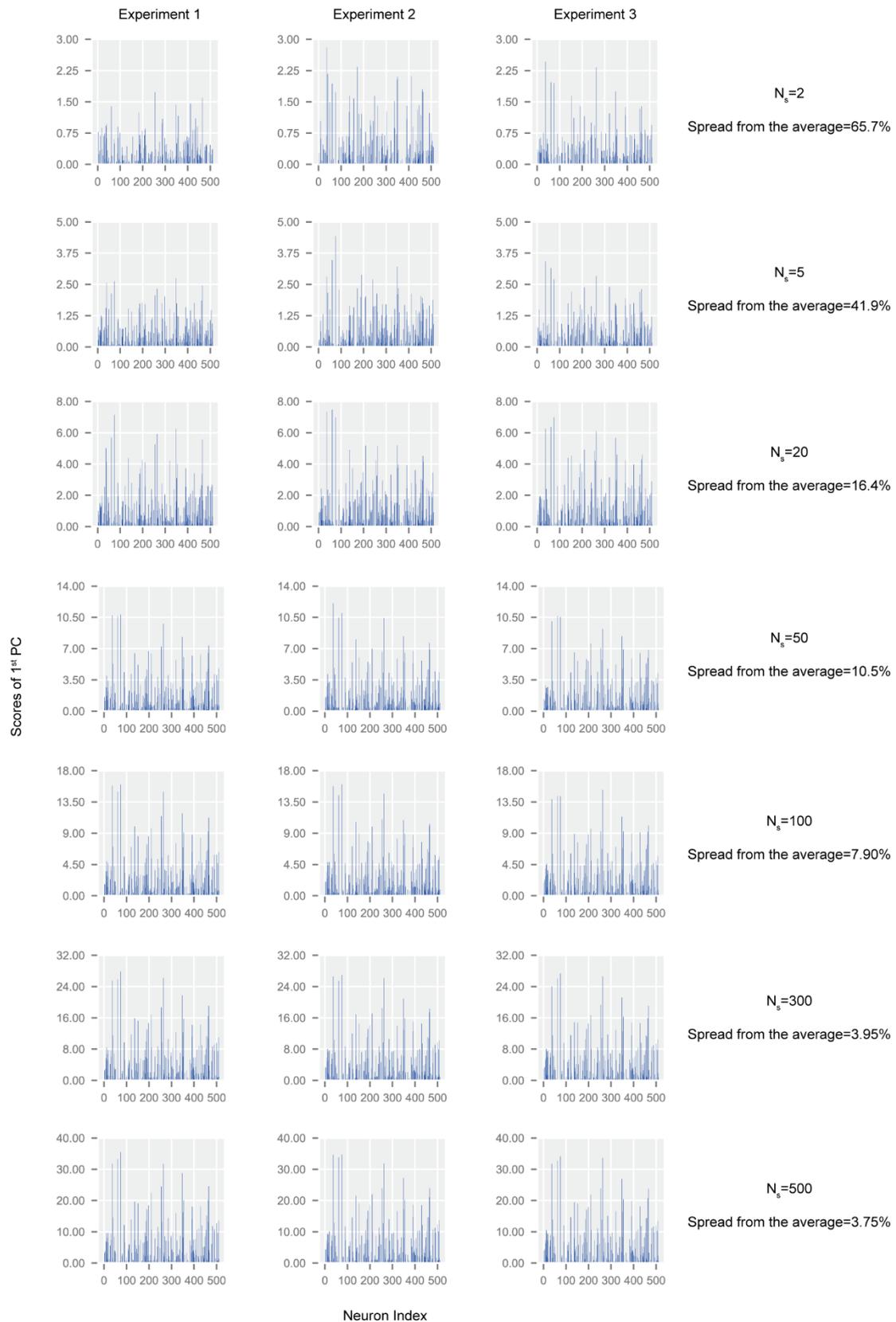

**Supplementary Fig. 3**. Scores of the 1st PC with different $N_s$ utilizing different random seeds to select different samples.

**S.2: Extension of the row-centered PCA**

The size of the feature map for each sample is a $C \times H \times W$ matrix, where $C$ is the number of channels, and $H$ and $W$ are the height and width of the feature map, respectively. In this work, the feature map is degenerated to a vector with the length of $C$ since the last layer is the global average pooling (GAP). In fact, the row-centered PCA can be extended to common feature maps generally. For $N_s$ feature maps extracted from $N_s$ samples, the data matrix is $N_s \times C \times H \times W$. The row-centered PCA for this data matrix can be conducted in the $N_s \times C$ submatrix for each point in the $H \times W$ submatrix. For the $H \times W$ times of the row-centered PCA, we uniformly retain $k$ principal components and obtain the reduced submatrix with the size of $k \times C$. Finally, we concrete the $k \times C$ and $H \times W$ submatrices to obtain the ultimate $k \times C \times H \times W$ matrix after the row-centered PCA.

Here, we conduct the row-centered PCA on the feature maps of each layer in the network to further explore the differences between different layers, and the information ratio of the 1st PC for each layer is illustrated in Supplementary Fig. 4. It is found that the information ratio of the 1st PC exhibits a trend of first decreasing and then increasing with the deepening of layers. Particularly, the last global average pooling (GAP) layer greatly promotes the information ratio of the 1st PC, which proves that it can realize dimension reduction, and preserve spatial information extracted by the previous convolutional layers and pooling layers, and thus concentrate the common traits. Meanwhile, considering that the information ratio of the 1st PC is closely related to the extraction of common traits, the convolutional layer and max pooling layer seem to contribute to concentrating the common traits, which may explain the powerful generalization ability of CNNs. The trend of information ratio of the 1st PC also provides solid evidence that the shallow layers of the CNN acquire simple texture features that have more common traits, the middle layers acquire local features that have fewer common traits, and the deep layers acquire overall category information, i.e., semantic information, that has more common traits.

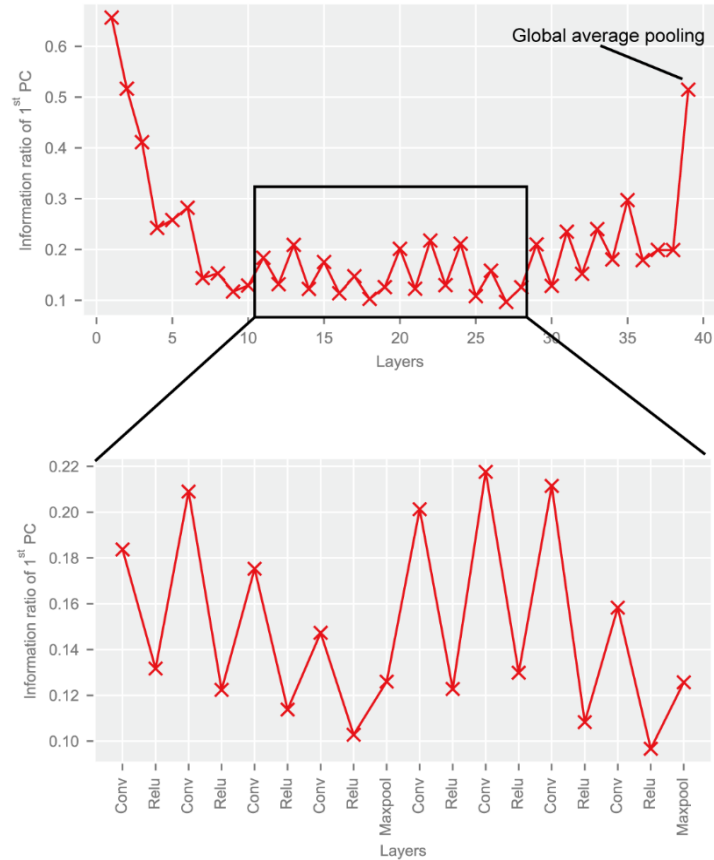

**Supplementary Fig. 4.** The information ratio of the 1st PC for each layer in the network (upper) and partial enlarged pictures (lower). In the upper part, the label of the x-axis is the layer index, and the last layer is the global average pooling (GAP) layer. In the lower part, the label of the x-axis is the class of the layer.

**S.3: Supplementary information for the statistical explanation of semantic space**

In this work, the distribution is close to part of the normal distribution in the semantic space of cats' eyes. Here, we provide the probability density distribution plots of the values of the weighted average activation $A_s$ for 3,000 samples of cats and dogs in different semantic spaces, including eyes, nose and legs, which are displayed in Supplementary Fig. 5. The respective quantile-quantile (q-q) plot, which is a graphical technique for determining if the given distribution is consistent with a normal distribution, is used. From the q-q plots, it is discovered that the $R^2$ are all above 0.97, which shows a strong correlation between the distribution and normal distribution in all semantic spaces. It is also worth noting that the distributions are consistent with the fitted normal distribution in the latter half, which proves that the distributions of semantic spaces are close to part of the normal distribution. Moreover, it is interesting to find that the distribution of the weighted average activation is actually close to part of the normal distribution and exhibits a slight difference. This is because the selected 3,000 samples cannot fully represent the concept of "cat" or "dog", which means that the dataset itself is imperfect. It is also worth mentioning that the semantic concepts here can be defined arbitrarily in a way that humans can understand, which enhances the interpretability of the CNN.

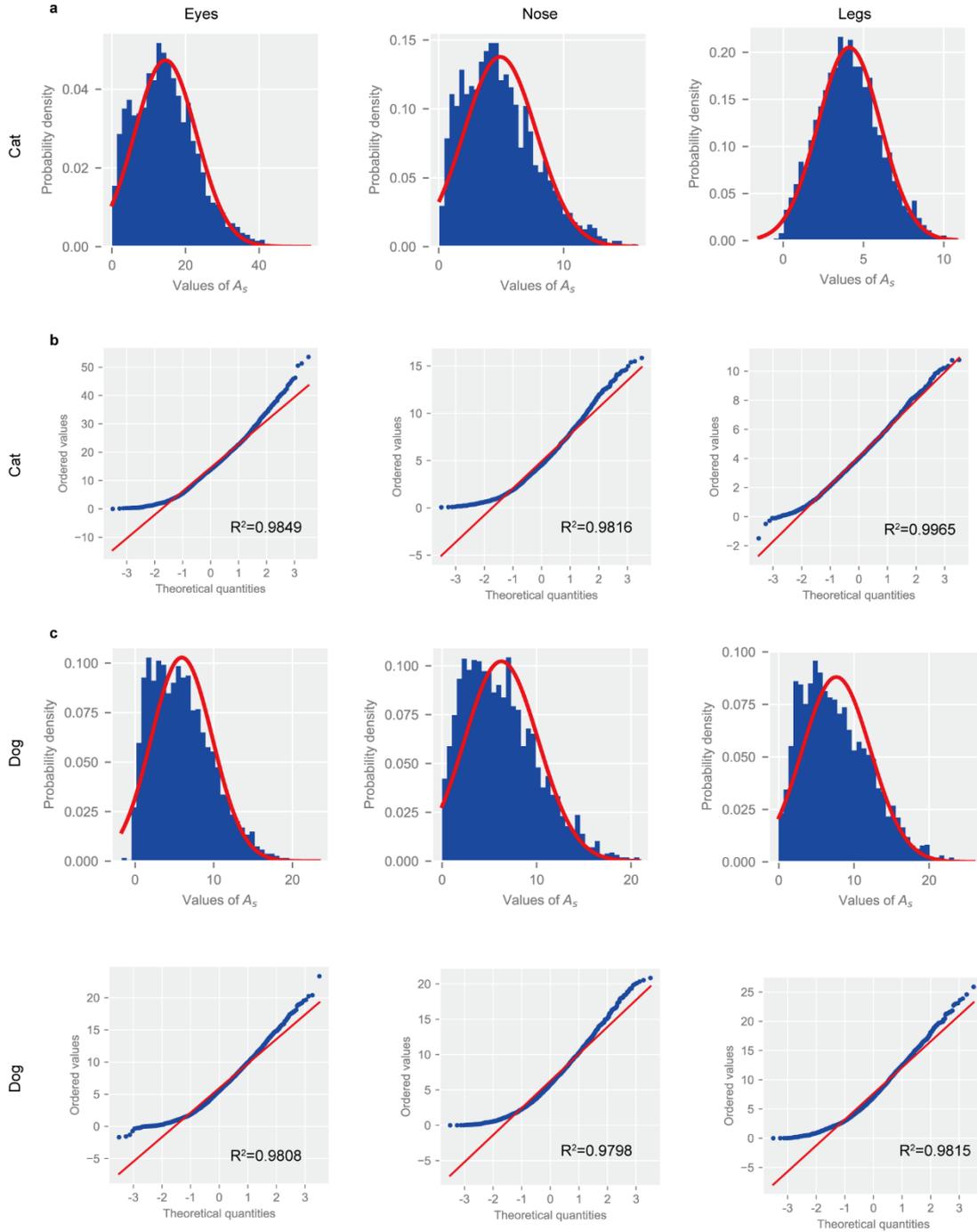

**Supplementary Fig. 5.** Probability density distribution plots of the values of the weighted average activation $A_s$ for 3,000 samples of cats (a) and dogs (c) in different semantic spaces, including eyes (left), nose (middle) and legs (right), where the red curves are the fitted normal distributions and the quantile-quantile (q-q) plots of the distribution for cats (b) and dogs (d), including eyes (left), nose (middle) and legs (right), and where the red lines are the quantile lines of the fitted normal distributions.

### S.4: Expansibility of the S-XAI to other structures of CNN
In this work, the VGG-19 network with a global average pooling (GAP) layer is employed mainly because the visualization of layers in the VGG-19 network has been proven to be effective and recognizable[1,2]. Here, we investigate the expansibility of our proposed S-XAI to other structures of CNN. We adopt the S-XAI to AlexNet with a GAP layer, and the results are shown in Supplementary Fig. 6. From the figure, it can be seen that S-XAI can also extract the semantic space in AlexNet, but accuracy is affected. First, the visualization of common traits extracted from AlexNet is so abstract that the semantic information is not explicit enough to be recognized like that in the VGG-19 network. In fact, in previous work, the visualization of AlexNet proves the existence of the phenomenon of distortion and abstraction[2]. However, the distorted silhouette still reveals some common traits that can differentiate cats and dogs. From the probability density distribution plot of the values of the weighted average activation and the q-q plot, it is discovered that although the distribution deviates more from the normal distribution compared with the VGG-19 network, the semantic probability can still represent the probability of the semantic concept well, which means that the semantic space is extracted successfully by S-XAI from AlexNet. In addition, it is found that the semantically sensitive neurons are similar for the cats' legs and dogs' legs in the semantic space, which makes it difficult to differentiate the cats' legs and dogs' legs in the semantic space. This suggests that AlexNet may not be highly sensitive to the legs of dogs and cats, which indicates that the ways of extracting semantic spaces may be different in different constructions of CNNs.

a

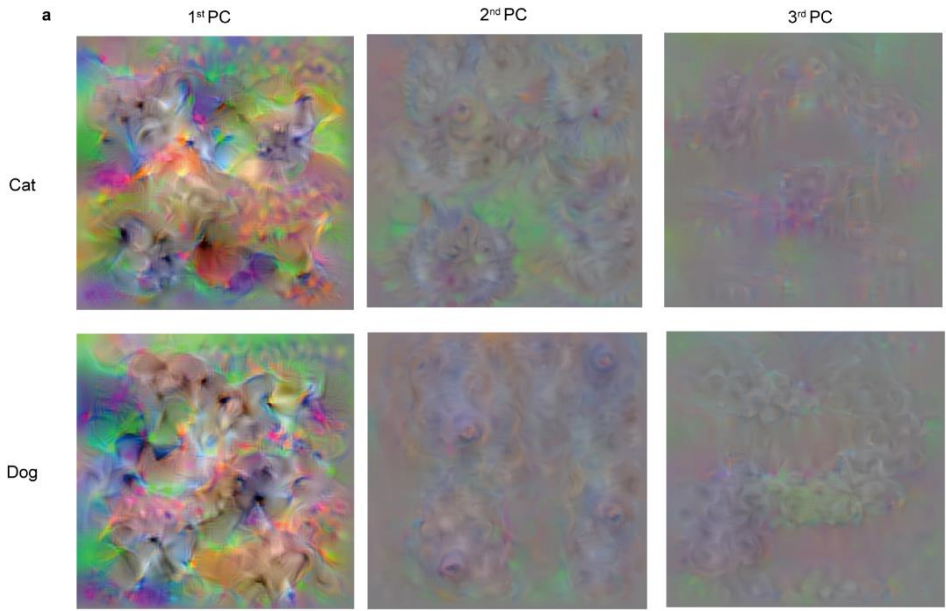

b

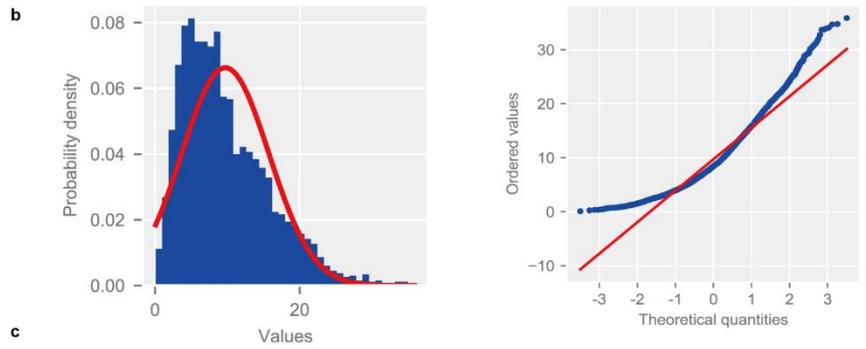

c

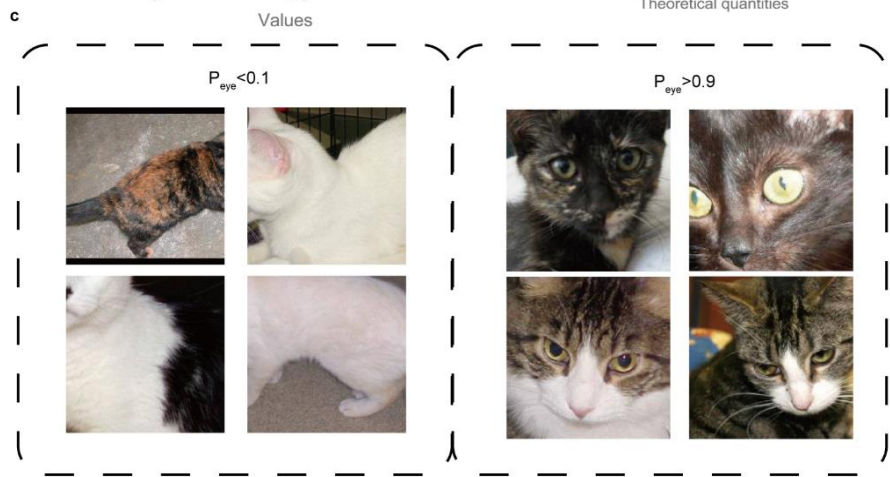

d

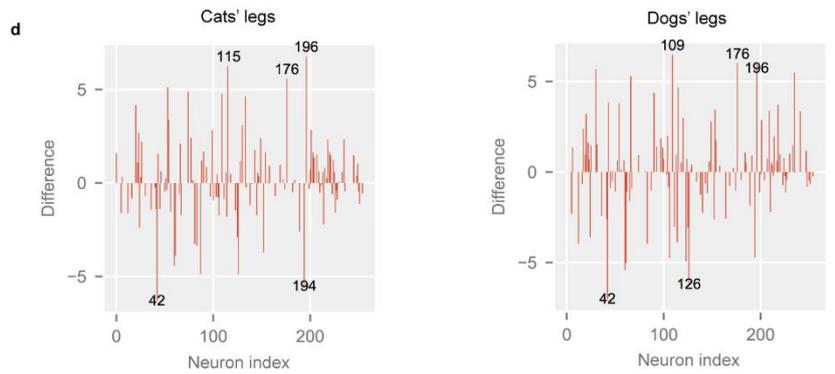

**Supplementary Fig. 6.** The results of S-XAI for AlexNet. **a,** Visualization of the 1st, 2nd, and 3rd PCs for cats and dogs with $N_s$=300, respectively. **b,** Probability density distribution plot of the values of the weighted average activation $A_s$ for 3,000 samples of cats in the semantic spaces of cats' eyes (left) where the red curve is the fitted normal distribution curve, and the quantile-quantile (q-q) plot of the distribution (right) where the red line is the quantile line of the fitted normal distribution. **c,** Samples located at the left ($P_{eye}$<0.1) and right ($P_{eye}$>0.9) ends of the distribution, respectively. **d,** The difference for the cats' legs (left) and dogs' legs (right), where the notations are the first five semantically sensitive neurons.

**S.5: Adversarial example identification**

Recently, many studies have been performed on adversarial samples, including L-BFGS[3], FCSM[4], PGD[5], etc., the purpose of which is to impose noise that cannot be discerned by human eyes on the samples to reduce the confidence of the neural network and induce incorrect assessments, which poses a challenge for the neural network to guarantee its safety[6]. Methods of defense against adversarial examples have also been extensively studied[7–10], the most notable of which is adversarial training that improves the robustness of DNNs against adversarial attacks by retraining the model on adversarial examples[4]. However, it is costly to retrain the neural network.

In this work, we attempt to understand the adversarial example and identify it from the perspective of semantic space. PGD is chosen to attack the CNN as an example. By comparing the probability density distribution of natural samples and adversarial samples in the semantic space, it is found that the location of adversarial samples is very close to the right end of the fitted normal distribution of natural samples, which means that the semantic probability of adversarial samples is very close to 1 or even greater than 1. The radar maps of the true sample and adversarial sample are provided in Supplementary Fig. 7, which reveals that the semantic probabilities of the adversarial examples are unusually large compared with the true samples. This is because in order to achieve a high attack success rate, the adversarial samples often contain excessive information about the incorrect label, which means that its activation in semantic space is much higher than that in natural samples. Therefore, the semantic space can identify adversarial examples to a certain extent, the results of which are shown in Supplementary Table 2. The criterion for identifying the adversarial example is simple, in which one of the semantic probabilities is higher than 0.99 or more than one of the semantic probabilities is higher than 0.9. From the results, the accuracy of identification of adversarial examples shows that the stronger is the adversarial attack, the higher is the success rate of identification by S-XAI; whereas, the weaker is the adversarial attack, the lower is the success rate of identification by S-XAI. This demonstrates that the identification of adversarial examples via the semantic space limits the strength of adversarial attacks, so that the attacker has to incur a greater cost to find suitable parameters to control the strength of adversarial attacks. Considering that the attack methods of adversarial samples emerge in an endless stream, the proposed method is not invulnerable. However, because of its low defense cost, it is highly suitable to be integrated into defense methods to improve the success rate of defense. Overall, the semantic space sheds light on the defense of the adversarial example, and better defense techniques may be inspired from semantic space in the future.

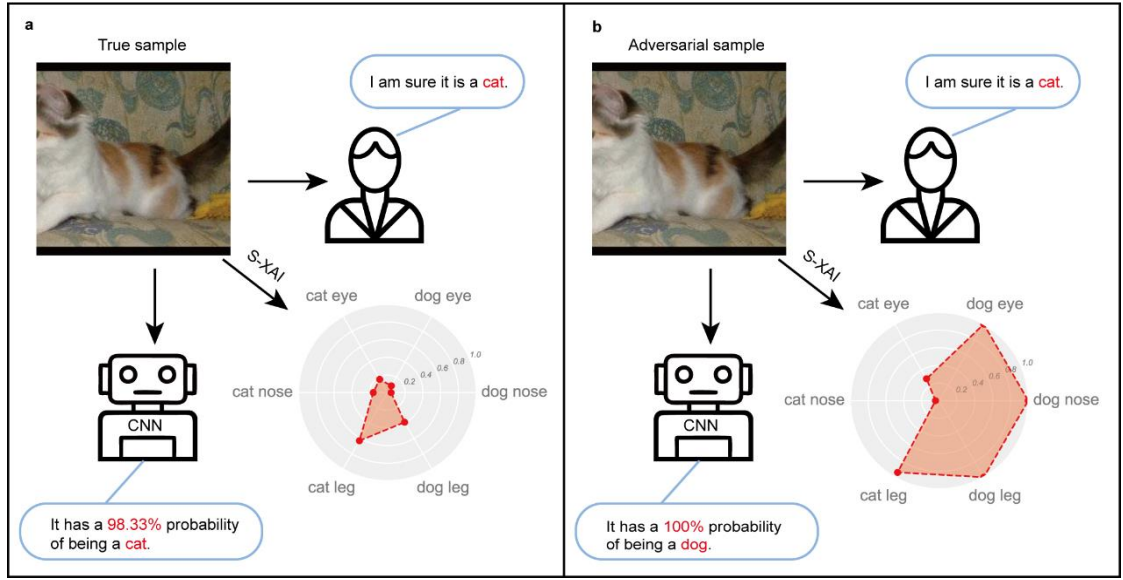

**Supplementary Fig. 7.** Assessments given by humans (right), CNN (bottom), and S-XAI (bottom right) when identifying the true sample (a) and the adversarial example (b).

**Supplementary Table 2.** The success rate of attack by PGD and the success rate of defense by S-XAI with different attack strengths.

| Parameters | $\varepsilon=0.05$ | $\varepsilon=0.01$ | $\varepsilon=0.005$ |
|---|---|---|---|
| Success rate of attack | 100 | 100 | 84 |
| Success rate of defense | 100 | 38 | 0 |

### S.6: Supplementary discussions

Although semantic spaces are successfully extracted in this work, certain shortcomings remain in the current research. First, the extraction of semantic space requires masking the semantic concepts in samples, which is completed manually. Although we have experimentally proven that only 100 masked images are sufficient to extract the semantic space well, manual annotation is still a major limitation, especially when faced with large-scale semantic space extraction with numerous semantic concepts and categories. Some other techniques, such as semantic segmentation or annotation-free techniques, for extracting object parts may assist to solve this problem. Second, numerous semantic concepts exist that are difficult to be masked in the image. For example, in this work, we use a set of typical semantic concepts to explain the CNN, including eyes, nose, and legs. However, certain infrequent semantic concepts, such as paws, beards, or tails, are difficult to include.

### S.7: Details of superpixel segmentation

The detailed process of superpixel segmentation is provided here. First, the image is converted into CIELAB color space $[l, a, b]^T$. Then, $k_s$ seed points (or cluster centers) are randomly initialized by sprinkling $k_s$ points on the image on average, which evenly fills the entire image. For the 3×3 area centered on each seed point, the gradient value of each pixel is calculated, and the point with the smallest value is selected as the new seed point, which aims to prevent the seed point from falling on the outline boundary. Afterwards, distance metrics for all pixels in the 2S*2S square area around the seed point are calculated, where $S=\sqrt{\frac{N_{sp}}{k_s}}$ and $N_{sp}$ is the number of image pixels. The distance metric $D$ between the pixels $i$ and $j$ is as follows:

$$D = \sqrt{\left(\frac{d_c}{N_c}\right)^2 + \left(\frac{d_s}{N_s}\right)^2}$$
$$d_c = \sqrt{(l_j - l_i)^2 + (a_j - a_i)^2 + (b_j - b_i)^2}.$$
$$d_s = \sqrt{(x_j - x_i)^2 + (y_j - y_i)^2}$$

where $d_c$ is the color metric; $d_s$ is the spatial metric; and $[x, y]^T$ is the pixel position. Since each pixel on the image may have several distance metrics calculated by different seed points, the seed point corresponding to the smallest distance metric is selected as its cluster center. This process will continue until the residual reaches the threshold set beforehand.

**References**

1. Mahendran, A. & Vedaldi, A. Understanding deep image representations by inverting them. in *Proceedings of the IEEE Computer Society Conference on Computer Vision and Pattern Recognition* vols 07-12-June-2015 (2015).
2. Lempitsky, V., Vedaldi, A. & Ulyanov, D. Deep Image Prior. in *Proceedings of the IEEE Computer Society Conference on Computer Vision and Pattern Recognition* 9446–9454 (2018). doi:10.1109/CVPR.2018.00984.
3. Szegedy, C. *et al.* Intriguing properties of neural networks. in *2nd International Conference on Learning Representations, ICLR 2014 - Conference Track Proceedings* 1–10 (2014).
4. Goodfellow, I. J., Shlens, J. & Szegedy, C. Explaining and harnessing adversarial examples. in *3rd International Conference on Learning Representations, ICLR 2015 - Conference Track Proceedings* (2015).
5. Madry, A., Makelov, A., Schmidt, L., Tsipras, D. & Vladu, A. Towards deep learning models resistant to adversarial attacks. in *6th International Conference on Learning Representations, ICLR 2018 - Conference Track Proceedings* (2018).
6. Chakraborty, A., Alam, M., Dey, V., Chattopadhyay, A. & Mukhopadhyay, D. A survey on adversarial attacks and defences. *CAAI Transactions on Intelligence Technology* vol. 6 25–45 (2021).
7. Meng, D. & Chen, H. MagNet: A Two-Pronged defense against adversarial examples. in *Proceedings of the ACM Conference on Computer and Communications Security* (2017). doi:10.1145/3133956.3134057.
8. Hendrycks, D. & Gimpel, K. Early methods for detecting adversarial images. in *5th International Conference on Learning Representations, ICLR 2017 - Workshop Track Proceedings* (2019).
9. Metzen, J. H., Genewein, T., Fischer, V. & Bischoff, B. On detecting adversarial perturbations. in *5th International Conference on Learning Representations, ICLR 2017 - Conference Track Proceedings* (2017).
10. Papernot, N., McDaniel, P., Wu, X., Jha, S. & Swami, A. Distillation as a Defense to Adversarial Perturbations Against Deep Neural Networks. in *Proceedings - 2016 IEEE Symposium on Security and Privacy, SP 2016* (2016). doi:10.1109/SP.2016.41.